\documentclass[10pt, lettersize,journal]{IEEEtran}
\usepackage{amsmath,amsfonts}
\usepackage{amssymb}
\usepackage{algorithmic}
\usepackage{array}
\usepackage[caption=false,font=normalsize,labelfont=rm,textfont=rm]{subfig}
\usepackage{textcomp}
\usepackage{stfloats}
\usepackage{url}
\usepackage{verbatim}
\usepackage{graphicx}
\usepackage[utf8]{inputenc}
\usepackage[ruled,vlined]{algorithm2e}
\def\BibTeX{{\rm B\kern-.05em{\sc i\kern-.025em b}\kern-.08em
		T\kern-.1667em\lower.7ex\hbox{E}\kern-.125emX}}
\usepackage{balance}
\usepackage{hyperref}
\usepackage{makecell}
\hypersetup{
	colorlinks = true, 
	linkcolor = blue,
	urlcolor = blue,
	citecolor = blue
}

\begin{document}
	%\pagewiselinenumbers
	% 设置行号在左侧
	%\linenumbers
	
	%\pagewiselinenumbers
	\captionsetup{font={small}}
	\title{Semantic-Aware Resource Management for C-V2X Platooning via Multi-Agent Reinforcement Learning}

	\author{Wenjun Zhang, Qiong Wu,~\IEEEmembership{Senior Member,~IEEE}, Pingyi Fan,~\IEEEmembership{Senior Member,~IEEE},
		\\Kezhi Wang,~\IEEEmembership{Senior Member,~IEEE}, Nan Cheng,~\IEEEmembership{Senior Member,~IEEE},
		\\Wen Chen,~\IEEEmembership{Senior Member,~IEEE}, and Khaled B. Letaief,~\IEEEmembership{Fellow,~IEEE}
		% <-this % stops a space
		\thanks{This work was supported in part by the National Natural Science Foundation of China under Grant 61701197 and Grant 62071296; in part by the National Key Research and Development Program of China under Grant 2021YFA1000500(4); in part by the National Key Project under Grant 2020YFB1807700; in part by Shanghai Kewei under Grant 22JC1404000; in part by the Research Grants Council under the Areas of Excellence Scheme under Grant AoE/E-601/22-R; and in part by the 111 Project under Grant B23008.
			
			Wenjun Zhang and Qiong Wu are with the School of Internet of Things Engineering, Jiangnan University, Wuxi 214122, China (e-mail: wenjunzhang@stu.jiangnan.edu.cn, qiongwu@jiangnan.edu.cn)
			
			Pingyi Fan is with the Department of Electronic Engineering, State Key Laboratory of Space Network and Communications, and the Beijing National Research Center for Information Science and Technology, Tsinghua University, Beijing 100084, China (e-mail: fpy@tsinghua.edu.cn).
			
			Kezhi Wang is with the Department of Computer Science, Brunel University, London, Middlesex UB8 3PH, UK (e-mail: Kezhi.Wang@brunel.ac.uk).
			
			Nan Cheng is with the State Key Laboratory of ISN and the School of Telecommunications Engineering, Xidian University, Xi'an 710071,China (e-mail: dr.nan.cheng@ieee.org).
			
			Wen Chen is with the Department of Electronic Engineering, Shanghai Jiao Tong University, Shanghai 200240, China (e-mail: wenchen@sjtu.edu.cn).

			Khaled B. Letaief is with the Department of Electrical and Computer Engineering, the Hong Kong University of Science and Technology, Hong Kong (e-mail: eekhaled@ust.hk).

		}% <-this % stops a space
		% <-this % stops a space
	}

	%% The paper headers
	%\markboth{IEEE Transactions on Vehicular Technology,~Vol.~XX, No.~XX, XXX~2015}%
	%{Shell \MakeLowercase{\textit{et al.}}: Bare Demo of IEEEtran.cls for IEEE Journals}

	% make the title area
	\maketitle
	
	\begin{abstract}
	Semantic communication transmits the extracted features of information rather than raw data, significantly reducing redundancy, which is crucial for addressing spectrum and energy challenges in 6G networks. In this paper, we introduce semantic communication into a cellular vehicle-to-everything (C-V2X)-based autonomous vehicle platoon system for the first time, aiming to achieve efficient management of communication resources in a dynamic environment. Firstly, we construct a mathematical model for semantic communication in platoon systems, in which the DeepSC model and MU-DeepSC model are used to semantically encode and decode unimodal and multi-modal data, respectively. Then, we propose the quality of experience (QoE) metric based on semantic similarity and semantic rate. Meanwhile, we consider the success rate of semantic information transmission (SRS) metric to ensure the fairness of channel resource allocation. Next, the optimization problem is posed with the aim of maximizing the QoE in vehicle-to-vehicle (V2V) links while improving SRS. To solve this mixed integer nonlinear programming problem (MINLP) and adapt to time-varying channel conditions, the paper proposes a distributed semantic-aware multi-modal resource allocation (SAMRA) algorithm based on multi-agent reinforcement learning (MARL), referred to as SAMRAMARL. The algorithm can dynamically allocate channels and power and determine semantic symbol length based on the contextual importance of the transmitted information, ensuring efficient resource utilization. Finally, extensive simulations have demonstrated that SAMRAMARL outperforms existing methods, achieving significant gains in QoE, SRS, and communication delay in C-V2X platooning scenarios.
	\end{abstract}
	
	% Note that keywords are not normally used for peerreview papers.
	
	\IEEEpeerreviewmaketitle
	
	\section{Introduction}
	\subsection{Background}
	\IEEEPARstart{W}{ith} the rapid development of intelligent transportation systems (ITS) \cite{r1}, ensuring safe and efficient transportation is increasingly crucial. A key component of ITS is platooning systems, where multiple autonomous vehicles travel closely together to enhance traffic flow and safety \cite{r3}. In these systems, a designated platoon leader (PL) manages formation, while platoon members (PMs) maintain coordinated speed and distance. Effective intra-platoon and inter-platoon communication (platoon-to-platoon or platoon-to-infrastructure) is essential for optimizing overall efficiency and safety.
	
	Integrating cellular vehicle-to-everything (C-V2X) communication is imperative for achieving effective communication. C-V2X supports two main types: vehicle-to-vehicle (V2V), which allows vehicles to share cooperative awareness messages (CAMs) for synchronized movement, and vehicle-to-infrastructure (V2I), enabling communication with base stations (BSs) for traffic and safety information \cite{r4}. These modalities facilitate timely responses to dynamic traffic conditions, enhancing the safety and efficiency of platooning systems. However, increased connectivity and communication demands in C-V2X platooning introduce complexities in managing network resources, which are critical for reliable communication in autonomous driving \cite{r5}. The frequency of information exchange among vehicles directly impacts their ability to react to obstacles, underscoring the need for efficient resource management \cite{r6,n5}.
	
	A promising approach to these challenges is semantic communication \cite{r7}, which focuses on transmitting extracted features (meaningful information) rather than raw data. This enhances data transmission efficiency in C-V2X systems by ensuring only relevant information is shared \cite{r8}, improving decision-making and reducing unnecessary data transfer.
	
	As task complexity rises, there is a shift from unimodal to multi-modal tasks that integrate various data types \cite{r9}. This integration fosters a richer understanding of the environment, enhancing network robustness and flexibility \cite{r10}. Moreover, traditional centralized resource management schemes often face inefficiencies and high signaling overhead due to their reliance on global information, particularly under dynamic channel conditions \cite{n1,r11,n3}.
	
	To address these challenges, we propose a distributed semantic-aware multi-modal resource allocation (SAMRA) framework leveraging multi-agent reinforcement learning (MARL), termed SAMRAMARL\footnote{The source code has been released at: https://github.com/qiongwu86/Semantic-Aware-Resource-Management-for-C-V2X-Platooning-via-Multi-Agent-Reinforcement-Learning}. By employing MARL, vehicles can make decentralized decisions based on local observations, reducing reliance on centralized control while enhancing scalability and adaptability to dynamic network conditions \cite{r13,n2}.
	
	\subsection{Related Work and Motivation}
	Resource allocation in platooning systems has garnered significant attention within intelligent transportation systems (ITS). Platooning involves multiple autonomous vehicles traveling in coordination, necessitating efficient communication between the platoon leader (PL), other platoon members (PMs), and infrastructure like BSs \cite{9878288}. Various studies have explored traditional methods to address these challenges.
	
	For instance, Guo \emph{et al.} \cite{RN1} proposed a joint optimization approach for LTE-V2V radio resource allocation and vehicle control parameters to enhance platoon stability. Hong \emph{et al.} \cite{RN2} designed a framework using relays and adaptive distributed model predictive control (DMPC) to improve safety in failure scenarios. Wang \emph{et al.} \cite{RN3} developed a two-step resource allocation strategy optimizing platoon formation and power control through branch and bound methods. Wen \emph{et al.} \cite{RN4} focused on optimizing inter-vehicle communication topology in LTE-V2V networks. However, the dynamic nature of channel conditions complicates effective resource management due to bias in estimating channel state information (CSI).
	
	To address these complexities, deep reinforcement learning (DRL) has emerged as a promising method for resource allocation in vehicular networks \cite{RN5, RN6, RN7, RN8}. For example, Liing \emph{et al.} \cite{RN5} utilized a multi-agent reinforcement learning (MARL) approach with deep Q-networks to enhance V2I capacity. Nasir \emph{et al.} \cite{RN6} proposed a model-free DRL-based power allocation scheme for wireless networks. Xu \emph{et al.} \cite{RN7} applied DRL for multi-objective resource allocation, focusing on transmission success and communication quality. Despite these advancements, distributed DRL still faces challenges in high-data-volume scenarios, leading to elevated signaling costs and delays, especially in dynamic environments \cite{10416925}.
	
	In response, unlike traditional methods that focus on raw data transmission, semantic communication prioritizes the meaningful features behind the conveyed information. This approach can enhance coordination and safety in platooning systems \cite{9679803}. For instance, Bourtsoulatze \emph{et al.} \cite{RN9} developed a method for efficient image transmission under low signal-to-noise ratio (SNR) conditions, while Huang \emph{et al.} \cite{RN10} used generative adversarial networks for semantic image compression.
	
	Moreover, integrating unimodal and multi-modal tasks in ITS is an important research area \cite{9642836}. Unimodal tasks handle specific data types, while multi-modal tasks provide a comprehensive understanding of the environment \cite{8884460}. However, existing studies often lack a unified resource management framework that leverages both semantic communication and the integration of unimodal and multi-modal tasks \cite{10431795}.
	
	Despite these advancements, there remains a gap in the literature. Existing approaches have not effectively combined distributed DRL, semantic communication, and the handling of both unimodal and multi-modal tasks into a cohesive resource management framework. This paper proposes a novel approach to optimize resource management in platooning systems by leveraging the strengths of semantic communication and distributed decision-making.
	
	\subsection{Contributions} 
	In this work, we present a distributed SAMRA algorithm based on MARL, termed SAMRAMARL, tailored for platooning systems. The primary contributions of our research are summarized as follows: 
	\begin{itemize} 
		\item[1)] We investigate the extraction of semantic and multi-modal information for C-V2X platooning systems, and redefine the metrics suitable for semantic and multi-modal data, as well as concept of quality of experience (QoE). 
		\item[2)] We formulate a joint optimization problem to maximize QoE and the success rate of semantic information transmission (SRS) in V2V links ,which is the first work to introduce semantic communication to address resource allocation challenges in platooning systems. 
		\item[3)] We design the SAMRAMARL algorithm, employing MARL to optimize various aspects of resource allocation, including channel assignment, power allocation, and the length of transmitted semantic symbols in both single-modal and multi-modal contexts. 
		\item[4)] Simulation results demonstrate that our SAMRAMARL algorithm significantly outperforms existing algorithms in terms of QoE, SRS, and communication delay. 
	\end{itemize}
	
	The rest of the paper is structured as follows: Section II introduces the system model and the formulated problem of maximizing QoE and SRS; Sections III present the proposed SAMRA algorithm; Section IV provides and discusses simulation results; and the paper is concluded in Section V.

	\section{System Model}\label{system}
\begin{figure*}
		\centering
		\includegraphics[width=0.8\textwidth]{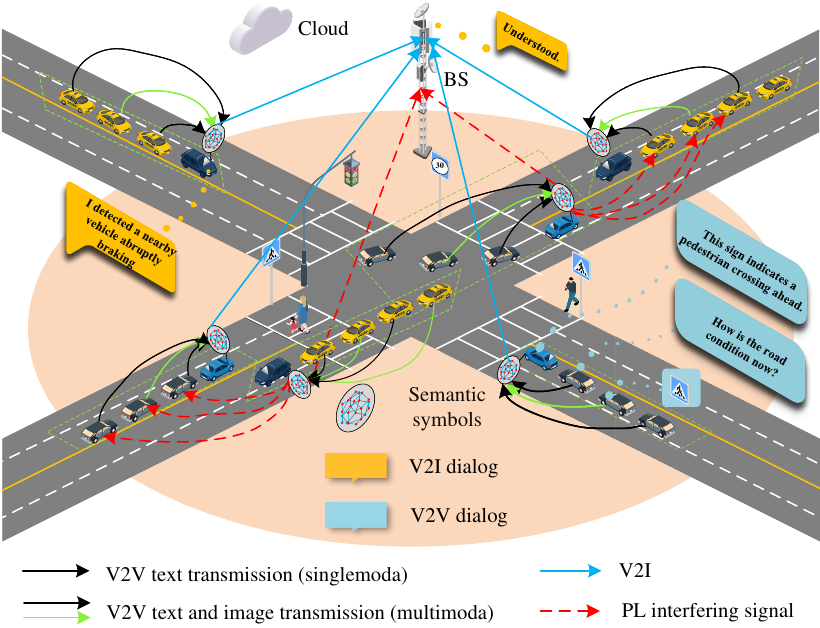}
		\caption{The system model}
		\label{fig1}
	\end{figure*}

	As shown in Fig. \ref{fig1}, we consider a C-V2X communication network that integrates a BS with $N$ platoons of autonomous vehicles. The set of platoons is denoted as $\mathcal{N} = \left\{ {1,2, \ldots ,N} \right\}$, $N \in \mathbb{N}$. Each platoon $n$ consists of ${M_{n}}$ vehicles, where $n \in \mathcal{N}$, and the first vehicle in each platoon is designated as the PL, which coordinates intra-platoon communication and handles inter-platoon communication with BS. 
	The set of vehicles in platoon $n$ is denoted as $\mathcal{M}_{n} = \left\{ {1, \ldots ,M_{n}} \right\}$, $M_{n} \in \mathbb{N}$. The communication system supports two types of communication: intra-platoon (V2V) and inter-platoon (V2I).
	In intra-platoon communication, the PL broadcasts semantic messages to its followers, and PMs transmit semantic symbols to the PL via V2V links. This guarantees the  platoon string stability and ensures that all the platoon members are aware of each other’s kinematics and decision-making. Inter-platoon communication occurs between the PL of each platoon and the BS over V2I links, where control and safety messages are exchanged to maintain coordination across platoons. Orthogonal frequency division multiplexing (OFDM) is utilized for communication. The total system bandwidth is divided into $K$ orthogonal subchannels with bandwidth $W$, and the set of subchannels is denoted as $\mathcal{K} = \left\{ {1, \ldots ,K} \right\}$, $K \in \mathbb{N}$. Both V2V and V2I communications take place over these subchannels, allowing efficient bandwidth management and interference mitigation.
	
	In the autonomous vehicle platoon, text and image data serve as the core modalities: textual data directly conveys explicit instructions or descriptions, while visual data enhances environmental awareness. By integrating these two modalities, the system can make appropriate decisions and maintain safety. Therefore, this system supports two communication modes: single-modal tasks and multi-modal tasks. For single-modal communication, the deep semantic communication (DeepSC) \footnote{DeepSC is not the only approach, but it is relatively mature and has been widely verified in terms of semantic compression and anti-noise performance.} model is employed, which is designed for textual data exchange \cite{9398576}. For multi-modal tasks, the multi-user deep semantic communication (MU-DeepSC) model is utilized, supporting both text and image data transmission \cite{9653664}. Note that when both models generate semantic symbols, they will prioritize according to the contextual importance. For example, the text of the emergency braking warning will be encoded as high-priority semantic symbols.
	In inter-platoon communication, vehicles primarily engage in single-modal communication, exchanging textual semantic information with the BS. The textual data is converted into semantic symbols using the DeepSC model. For instance, if a vehicle detects a sudden braking event, it could transmit the corresponding semantic information as a warning message to the BS, which then disseminates this information to the relevant platoons.
	In intra-platoon communication, the use of multi-modal tasks is determined by the number of PMs. If the number of PMs is even, vehicles engage in multi-modal communication, where each pair of vehicles employs the MU-DeepSC model to exchange both textual and image-based semantic information. If the number of PMs is odd, multi-modal communication is applied to pairs of vehicles, while the remaining vehicle communicates use single-modal transmission.
	For example, within a platoon, one vehicle captures an unfamiliar traffic sign while another vehicle inputs a text query to inquire about the sign's meaning. Both pieces of semantic information are sent to the PL, who, after processing, provides an accurate interpretation of the sign, like ``This sign indicates a pedestrian crossing ahead".
	
	The following subsections introduce the principles of the two semantic communication models, DeepSC and MU-DeepSC, followed by an analysis of key performance metrics, including channel gain, signal-to-interference-plus-noise ratio (SINR), and interference. Finally, we present a new semantic QoE model to evaluate overall performance in this vehicular network.	
	
	\subsection{Semantic Communication Model}
	\subsubsection{Semantic Transceivers}
	The transmitted data in both V2V and V2I communication undergoes semantic encoding and channel encoding before being sent over the wireless channel.
	In single-modal communication, vehicles exchange textual semantic information using the DeepSC model \cite{9398576}.
	Let ${S_{T}}$ represent the input sentence, which is first embedded into a numerical vector. The semantic encoding and channel encoding are performed together in the following manner:
	\begin{equation}\label{eq1}
		{X_{T}} = c{h_{{\beta _T}}}(s{e_{{\alpha _T}}}({S_{T}})),
	\end{equation}
	where $s{e_{{\alpha _T}}}( \cdot )$ represents the semantic encoding function utilizing a Bi-directional long short-term memory (Bi-LSTM) network, parameterized by ${\alpha _T}$. This encoder extracts meaningful semantic features from the input text. 
	Subsequently, $c{h_{{\beta _T}}}\left(  \cdot  \right)$, the channel encoder, compresses and maps the semantic features into transmitted symbols suitable for communication over the physical channel. The combined operation ensures that only the essential, extracted features full of semantic information is transmitted, optimizing resource usage in the V2X environment.
	
    For multi-modal communication, vehicles exchange both textual and image semantic information, which is processed using the MU-DeepSC model. The model handles both image and text inputs simultaneously, ensuring effective transmission of diverse data types.  Consider ${S_{I}}$ as the input image and ${S_{T}}$ as the input text. The semantic and channel encoding for both modalities are performed in a coordinated manner.
	For the image input, the semantic encoding captures essential features from the image, and the channel encoding prepares the information for transmission: 
	\begin{equation}\label{eq2}
		{X_{I}} = c{h_{{\beta _I}}}(s{e_{{\alpha _I}}}({S_{I}})),
	\end{equation}
	where $s{e_{{\alpha _I}}}( \cdot )$ is the image semantic encoder (based on a deep convolutional neural network, such as ResNet-101), and $c{h_{{\beta _I}}}( \cdot )$ is the channel encoder responsible for translating the extracted semantic features into symbols. The text input follows a similar encoding process as described for the DeepSC model. Both the image-transmitted symbols ${X_{I}}$ and the text-transmitted symbols ${X_{T}}$ are transmitted together or sequentially, depending on the network’s transmission protocol. This approach ensures that both image and textual semantic information are conveyed efficiently, maintaining the integrity of multi-modal data while optimizing bandwidth and reducing redundancy.
	\subsubsection{Transmission Model}
	After semantic and channel encoding, the transmitted symbols ${X_{T}}$ (textual) and ${X_{I}}$ (image) are sent over the wireless channels in the platooning system. We consider a distributed resource allocation system where the platoons choose resources autonomously. However, hidden node problems or decision delays may lead to subchannel conflicts and hence interference. For example, two distant platoons choose the same sub-channel at the same time because they cannot sense each other. Alternatively, there may be a transient overlap in the sub-channel selection decisions of different platoons in dynamic environments. 
	
	The system consists of two types of communication: intra-platoon communication (V2V) and inter-platoon communication (V2I), both subject to channel effects and interference. To present the wireless transmission model with a clearer distinction between V2V (intra-platoon) and V2I (inter-platoon) communication, we define the binary decision variable ${\rho _{n,k}} \in \{ 0,1\}$. This variable indicates the choice of communication type for platoon $n$ over subchannel $k \in \mathcal{K}$. Specifically, when ${\rho _{n,k}} = 1$, PL uses subchannel $k$ for V2V communication to broadcast messages within its own platoon; otherwise, the PL uses subchannel $k$ for V2I communication to communicate with BS.
	After semantic and channel encoding, the transmitted symbols ${X_{T,n}}$ (text) and ${X_{I,n}}$ (image) from the PL of platoon $n$ are transmitted over the wireless channel in the platooning system. The received signals, interference, and SINR are modeled separately for V2V and V2I communication, taking into account the choice of ${\rho _{n,k}}$.
	
	In V2V communication, vehicles exchange information through direct communication links within the same platoon. The received signals at vehicle $m \in \mathcal{M}_{n}$ within platoon $n$ over subchannel $k$ are expressed as follows:
	for text transmission: 
	\begin{equation}\label{eq3}
		Y_{T, n, m}[k] = \rho_{n,k} h_{n, m}[k] X_{T, n}[k] + I_{n, m}[k] + {\chi _{n,m}},
	\end{equation} 
	for image transmission: 
	\begin{equation}\label{eq40}
		Y_{I, n, m}[k] = \rho_{n,k} h_{n, m}[k] X_{I, n}[k] + I_{n, m}[k] + {\chi _{n,m}},
	\end{equation} 
	where $\rho_{n,k} h_{n, m}[k]$ is the channel gain between the PL of platoon $n$ and vehicle $m$ over subchannel $k$, $I_{n, m}[k]$ represents the interference from other platoons transmitting on the same subchannel, and $\chi_{n,m}$ denotes the additive white Gaussian noise (AWGN), respectively. $I_{n, m}[k]$ is modeled as follows:
	\begin{equation}\label{eq40}
		{I_{n,m}}[k] = \sum\limits_{n' \ne n} {{\rho _{n',k}}{\beta _{n',k}}} {p_{n'}}[k]{h_{n',m}}[k],
	\end{equation}
	where ${{\beta _{n',k}}}$ indicates whether subchannel $k$ is allocated to platoon ${n'}$, \(\rho_{n',k}\) indicates that subchannel \(k\) is used for V2V communication or V2I communication by other platoons \(n'\), and ${{p}_{n'}}[k]$ is the transmit power, respectively.
	The corresponding signal-to-interference-plus-noise ratio (SINR) for text and image transmissions are given by: for text transmission: 
	\begin{equation}\label{eq5}
		\text{SINR}_{T,n,m}[k] = \frac{\rho_{n,k} p_{T,n}[k] h_{n,m}[k]}{I_{n, m}[k] + \sigma^2},
	\end{equation}
	for image transmission: 
	\begin{equation}\label{eq6}
		\text{SINR}_{I,n,m}[k] = \frac{\rho_{n,k} p_{I,n}[k] h_{n,m}[k]}{I_{n, m}[k] + \sigma^2},
	\end{equation}
	where ${{p_{T,n}}[k]}$ and ${{p_{I,n}}[k]}$ denote the transmit power levels allocated to text and image transmissions  over subchannel $k$, respectively, ${\sigma ^2}$ is the noise power.
	
	In V2I communication, the PL communicates with the BS using subchannel \(k\). This mode is applied when \(\rho_{n,k} = 0\). The received signal at the BS from the PL of platoon \(n\) over subchannel \(k\) is given by:
	\begin{equation}\label{eq3}
		Y_{T, \text{BS}}[k] = (1 - \rho_{n,k}) h_{n, \text{BS}}[k] X_{T, n}[k] + I_{T, \text{BS}}[k] + \chi _{T, \text{BS}},
	\end{equation} 
	where $(1 - \rho_{n,k}) h_{n, BS}[k]$ is the channel gain between the PL of platoon $n$ and the BS over subchannel $k$ when \(\rho_{n,k} = 0\). 
	\(I_{T, \text{BS}}[k]\) denotes the interference from other platoons using the same subchannel for V2I communication, and $\chi _{T, \text{BS}}$ represents the noise at the BS. \(I_{T, \text{BS}}[k]\) is modeled as follows:
	\begin{equation}\label{eq40}
		{I_{T,BS}}[k] = \sum\limits_{n' \ne n} {{(1 - \rho _{n',k})}{\beta _{n',k}}} {p_{T,n'}}[k]{h_{n',BS}}[k],
	\end{equation}
	The SINR for inter-platoon communication is: 
	\begin{equation}\label{eq40}
		\text{SINR}_{T,n,\text{BS}}[k] = \frac{(1 - \rho_{n,k}) p_{T,n}[k] h_{n, \text{BS}}[k]}{I_{T, \text{BS}}[k] + \sigma^2}.
	\end{equation} 

	\subsubsection{Receiver Model}
	The receiver model recovers the transmitted semantic symbols through a combination of channel decoding and semantic decoding, depending on whether the data is single-modal or multi-modal. 
	For single-modal text transmission, the received signal $Y_{T, n, m}[k]$ is decoded using a channel decoder followed by a semantic decoder: 
	\begin{equation}\label{eq_rx_single} 
		{{\hat S}_T} = s{d_{{\alpha _T}}}(c{d_{{\beta _T}}}({Y_{T,n,m}}[k])), 
	\end{equation} 
	where $c{d_{{\beta _T}}}( \cdot )$ is the channel decoder, and $s{d_{{\alpha _T}}}( \cdot )$ represents the semantic decoder that recovers the semantic meaning from the transmitted symbols.
	
	For multi-modal communication, the received signals $Y_{T, n, m}[k]$ and $Y_{I, n, m}[k]$ for text and image are decoded independently through their respective decoders: \begin{equation}\label{eq_rx_multi} \begin{array}{l}
			{{\hat S}_T} = s{d_{{\alpha _T}}}(c{d_{{\beta _T}}}({Y_{T,n,m}}[k]))\\
			{{\hat S}_I} = s{d_{{\alpha _I}}}(c{d_{{\beta _I}}}({Y_{I,n,m}}[k]))
		\end{array},
	\end{equation} 
	where $s{d_{{\alpha _I}}}( \cdot )$ is image semantic decoder, respectively. These decoded symbols are then fused to jointly infer the final semantic message.
	
	\subsection{Novel Metrics for Resource Management in platooning Networks}
	In vehicular platooning networks, effective resource management is crucial for maintaining efficient communication, particularly for autonomous driving applications. To optimize resource allocation and improve overall system performance, we propose novel metrics (semantic similarity and semantic rate) and quality of experience (QoE). These metrics aim to assess the efficiency of semantic communication by directly quantifying the amount of meaningful information being transmitted, going beyond traditional metrics that focus mainly on bit-level accuracy.
	
	Semantic similarity is denoted as the extent to which the semantic information recovered at the receiver corresponds to the original information. Semantic rate is defined as the amount of semantic information transmitted per unit time. Both metrics are essential in vehicular networks, where communication must be both accurate and timely to ensure safe and coordinated actions within a platoon. Semantic rate is modeled based on semantic entropy. Unlike Shannon entropy, which measures the uncertainty of a source in terms of bits, semantic entropy quantifies the amount of meaningful information that a source carries with respect to a specific task. Calculating exact semantic entropy can be intractable for complex tasks. As such, we adopt deep learning (DL) models, such as DeepSC for single-modal and MU-DeepSC for multi-modal tasks, to approximate the semantic entropy\cite{10001594}. This approximation, expressed in semantic units (suts), serves as the basis for optimizing resource allocation in platooning networks, where the goal is to transmit only the most critical feature information.
	
	However, changes in the length of transmitted semantic symbols have opposite effects on semantic similarity and rate. As the length of transmitted semantic symbols decreases, the rate increases, but the similarity decreases. Therefore, we introduce a QoE model that assesses the overall performance of semantic communication systems by combining semantic accuracy and semantic rate. The QoE model proposed allows users to independently adjust their preferences for rate and accuracy, thus explicitly managing the trade-off between them.
	
	\subsubsection{Semantic Similarity}
	For single-modal text communication, we employ semantic similarity ${\xi _{SM}}$) as a performance metric to evaluate the accuracy of semantic transmission \cite{9398576}. This metric quantifies how closely the received message resembles the transmitted message in terms of its meaning. The semantic similarity can be expressed as:
	\begin{equation}\label{eq13}
		{\xi_{SM}} = {\Psi_{SM}}\left(u_m^T, \text{SINR}_{T,n,m/BS}[k]\right),
	\end{equation}
	where the function ${\Psi _{SM}}$ maps these inputs to the semantic similarity score, and $0 \le {\xi _{SM}} \le 1$, with ${\xi _{SM}} = 1$ indicating the highest similarity between the two sentences and ${\xi_{SM}} = 0$ implying no similarity. The term ${u_m^T}$ represents the average number of symbols required to encode each word as a semantic symbol (i.e., the semantic symbol length) by vehicle $m$, in units of suts/word. $SIN{R_{T,n,m/BS}}\left[ k \right]$ denotes the SINR for text communication received at vehicle $m$ in platoon $n$ or at the BS over subchannel $k$.
	
	For multi-modal communication, such as the fusion of text and image data, the semantic similarity $\xi_{MM}$ assesses joint performance of them. This metric evaluates how well the combined text and image information matches the expected outcome. The semantic similarity in this context is given by:
	\begin{equation}\label{eq13}
		{\xi_{MM}} = {\Psi_{MM}}\left(u_m^T, u_m^I, \text{SINR}_{T,n,m}[k], \text{SINR}_{I,n,m}[k]\right),
	\end{equation}
	where the function ${\Psi _{MM}}$ maps these inputs to the overall semantic similarity score. ${u_m^T}$ and ${u_m^I}$ denote the semantic symbol length for text and image, respectively, by vehicle $m$.
	
	\subsubsection{Semantic Rate}
	The semantic rate measures the volume of meaningful feature information transmitted by a vehicle per second, reflecting the effectiveness of data communication in vehicular networks. Unlike traditional data rates, that focus on raw data transfer, the semantic rate emphasizes the transmission of information that is relevant for understanding and decision-making processes in specific tasks. 
	The semantic rate ${\phi _n}$ is defined as:
	\begin{equation}\label{eq66}
		{\phi _m} = \frac{{W{{\tilde H}_{SM}}}}{{u_m^T}},
	\end{equation}
	where $W$ represents the bandwidth allocated to the communication channel, ${{{\tilde H}_{SM}}}$ represents the approximate semantic entropy of the textual data. 
	Semantic entropy is a critical concept that measures the amount of meaningful content within a message that is necessary for a specific task, rather than merely quantifying the raw data. Semantic entropy $H(X;Y)$ is defined as the minimum expected number of semantic symbols required to accurately predict or infer a task $Y$ from the data $X$ \cite{chattopadhyay2021quantifying}.
	\begin{equation}
		\begin{array}{l}
			H(X;Y) \triangleq \mathop {\min }\limits_{{{\rm E}_s}} {\rm E}(\dim (Cod{e^{{{\rm E}_s}}}(X)))\\
			s.t.\qquad{\rm{   }}P(Y|Cod{e^{{{\rm E}_s}}}(X)){\rm{ }} \approx P(Y|X),
		\end{array} 
	\end{equation}
	where $Cod{e^{{{\rm E}_s}}}(X)$ represents the semantic symbols extracted from $X$ using a semantic encoder ${{{\rm E}_s}}$, and $P(Y|Cod{e^{{{\rm E}_s}}}(X))$ approximates the conditional probability of predicting $Y$ given $X$. This definition emphasizes that the goal of semantic entropy is to retain only the most relevant information for a particular task, optimizing communication efficiency, especially in dynamic and resource-constrained environments like vehicular networks.
	Computing an exact measure of semantic entropy is challenging due to the difficulty in determining an optimal semantic encoder. To address this, we use a deep learning-based approximation denoted as $\tilde H(X;Y)$. 
	\begin{equation}
		\begin{array}{l}
			\tilde H(X;Y) \triangleq \mathop {\min }\limits_{s{e_\alpha }} {\rm E}(\dim (Cod{e^{s{e_\alpha }}}(X)))\\
			s.t.\qquad \left \lvert P(Y|Cod{e^{s{e_\alpha }}}(X)){\rm{ }} - P(Y|X) \right \lvert \textless \varepsilon,
		\end{array} 
	\end{equation}
	where the constraint indicates that the gap between $P(Y|Cod{e^{s{e_\alpha }}}(X))$ and $P(Y|X)$ can not exceed $\varepsilon$.
	This approach allows us to estimate the semantic content of a message by employing a well-trained semantic encoder ${s{e_\alpha }}$, capturing the key elements of the original data $X$ that are relevant for predicting the task $Y$.
	
	In the context of multi-modal communication, where both text and image data may be transmitted simultaneously, the semantic rate must be adapted for each data type to ensure effective transmission. The semantic rates for text and image data are given by:
	\begin{equation}\label{eq13}
		{\phi _{{m_{_T}}}} = \frac{{W\tilde H_{MM}^T}}{{u_m^T}},{\rm{   }}{\phi _{{m_{_I}}}} = \frac{{W\tilde H_{MM}^I}}{{u_m^I}},
	\end{equation}
	where ${\tilde H_{MM}^T}$ denotes the approximate semantic entropy for the multi-modal data of text and ${\tilde H_{MM}^I}$ is for image.
	
	In platooning networks, the PL must maintain timely communication with the BS to exchange critical intersection safety messages. Additionally, the PL and PMs engage in real-time interactions to coordinate movement and ensure the platoon's integrity. The V2V links play a crucial role in reliably disseminating safety-critical semantic messages that are generated periodically with varying frequencies depending on vehicle mobility and driving conditions.
	To ensure that these safety-critical semantic messages are delivered on time, we model the delivery rate of semantic packets of size \(B_s\) (in suts) within a time budget \(\Delta T\). The requirement is defined as:
	\begin{equation}\label{eq16}
		\sum_{t=1}^T \sum_{m=1}^{M_{n}} {{\beta _{n,k}}} \phi_{k}[m, t] \geq \frac{B_s}{\Delta T}, \quad \forall m \in \mathcal{M}_{n}, \forall k \in \mathcal{K},
	\end{equation}
	where \(B_s\) denotes the size of the periodically generated V2V payload in semantic units (suts), \(\Delta T\) is the channel coherence time, and \(\phi_{k}[m, t]\) represents the semantic rate of the \(k\)-th V2V link for vehicle \(m\) at time slot \(t\). The variable \({{\beta _{n,k}}}\) indicates whether subchannel \(k\) is assigned to platoon \(n\) (i.e., \({{\beta _{n,k}}} = 1\) if assigned, $0$ otherwise). This condition ensures that the required amount of semantic information is delivered within the time constraints, thus enabling safe and synchronized movements within the platoon.
	
	\subsubsection{QoE}
	Different vehicles within a platoon may prioritize these different metrics depending on their specific application needs demand. For instance, some vehicles might emphasize high semantic accuracy and tolerate some delays, while others may prioritize a higher semantic transmission rate even if it slightly compromises accuracy. To capture these varying preferences, we propose a QoE model that integrates both semantic accuracy and semantic rate. The QoE for vehicle \(m\) in platoon \(n\) is defined as:
	\begin{equation}\label{eq17}
		\begin{aligned}
			\text{QoE}_m^n &= \sum_{m \in \mathcal{M}_{n}} \left[\omega_m \text{Score}_R(\phi_m) + (1 - \omega_m) \text{Score}_A(\xi_m)\right] \\
			&= \sum_{m \in \mathcal{M}_{n}} \left[\frac{\omega_m}{1 + e^{\gamma (\phi_{target} - \phi_m)}} + \frac{1 - \omega_m}{1 + e^{\delta (\xi_{target} - \xi_m)}}\right],
		\end{aligned}
	\end{equation}
	where \(\text{QoE}_m^n\) represents the quality of experience for vehicle \(m\) in platoon \(n\). \(\omega_m\) is the weight assigned to the semantic rate for vehicle \(m\), while \(1 - \omega_m\) represents the weight for semantic accuracy. \(\phi_{target}\) and \(\xi_{target}\) are the target levels of semantic rate and accuracy considered optimal for the platoon's safe operation, and \(\gamma\) and \(\delta\) are parameters that determine the sensitivity of the scoring functions to deviations from these target values.
	
	\subsection{Optimization Problem}
	In this section, we formulate the semantic-aware resource allocation problem with the dual objectives of maximizing QoE for all vehicles within the platooning network and enhancing SRS of V2V. If only QoE is optimized, high QoE may be traded off by increasing the semantic symbol length to improve semantic similarity, but this will take up more channel resources and crowd out emergency message transmission opportunities for other vehicles, thus failing to transmit high-priority safety messages (e.g., emergency braking warnings) successfully within the strict time limit (\(\Delta T\)). In the autonomous vehicle platoon, the loss of latency-sensitive messages may lead to serious accidents. The proposed joint optimization framework bridges the above drawbacks.
	
	In the optimization problem, we have the variables such as channel assignment, power allocation, and the selection of the length of transmitted semantic symbols for both text and image data. The problem can be formally defined as follows:
	\begin{subequations}\label{P1} 
		\begin{equation}\label{P1a}\begin{array}{l}
				\mathop {{\rm{Maximize}}}\limits_{{\beta _n},{\rho _n},{p_n},{u_m^T},{u_m^I}} \\
				\sum\limits_{n \in \mathcal{N}} {\sum\limits_{m \in \mathcal{M}_{n}} {\left[ {QoE_m^n{\rm{ + }}\lambda \Pr \left( {\sum\limits_{t = 1}^T {\sum\limits_{m = 1}^{M_{n}} {{\beta _{n,k}}} } \phi _k[m,t] \ge \frac{{{B_s}}}{{\Delta T}}} \right)} \right]} } 
			\end{array} 
		\end{equation} 
		\begin{equation}\label{P1b} \text{s.t.} \quad \beta_{n,k} \in {0,1}, \quad \forall n \in \mathcal N, \forall k \in \mathcal K, 
		\end{equation} 
		\begin{equation}\label{P1c} \sum_{n \in \mathcal N} \beta_{n,k} \leq 1, \quad \forall k \in \mathcal K, 
		\end{equation} 
		\begin{equation}\label{P1d} \sum_{k \in \mathcal K} \beta_{n,k} \leq 1, \quad \forall n \in \mathcal N, 
		\end{equation} 
		\begin{equation}\label{P1e} 0 \leq u_m^T \leq u_{m,\max}^T, u_m^T \in \mathbb{N}, \quad \forall m \in \mathcal{M}_{n},
		\end{equation} 
		\begin{equation}\label{P1f} 0 \leq u_m^I \leq u_{m,\max}^I, u_m^I \in \mathbb{N}, \quad \forall m \in \mathcal{M}_{n}, 
		\end{equation} 
		\begin{equation}\label{P1g} 0 \leq p_m \leq p_{\max}, \quad \forall m \in \mathcal{M}_{n}, \forall k \in \mathcal{K}, 
		\end{equation} 
		\begin{equation}\label{P1h} Scor{e_{R,m}}, Scor{e_{A,m}} \geq {G_{th}}, \quad \forall m \in \mathcal{M}_{n}, \forall k \in \mathcal K, 
		\end{equation} 
		\begin{equation}\label{P1i} 0 \leq \sum_{k=1}^K \sum_{n=1}^N B_s \leq B_s^{\max}. 
		\end{equation} 
	\end{subequations}
	
	The objective function in (\ref{P1a}) aims to maximize the combined QoE of all vehicles and SRS within the time budget ${\Delta T}$. Here, $\lambda $ is a weight factor that balances the importance of QoE and timely delivery of safety-critical messages. The term $\Pr \left(  \cdot  \right)$ represents the probability the semantic packets are successfully transmitted over the V2V link in time, denoted as SRS.
	Constraints (\ref{P1b}) ensure that the channel allocation variable ${\beta _{n,k}}$ to be binary, meaning each subchannel $k$ is either assigned to platoon $n$ or not. (\ref{P1c}) imposes each subchannel to be allocated to at most one platoon, while (\ref{P1d}) ensures each platoon can be assigned at most one subchannel. 
	Constraints (\ref{P1e}) and (\ref{P1f}) define the range for the number of semantic symbols transmitted for both text ${u_m^T}$ and image ${u_m^I}$ data.
	Constraint (\ref{P1g}) constrains the transmission power ${p_m}$ for each vehicle, within system boundaries a space. Constraints (\ref{P1h}) ensure that the semantic rate and accuracy scores meet the minimum threshold ${G_{th}}$, guaranteeing acceptable QoE for all users. Finally, constraint (\ref{P1i}) ensures that the total semantic information size ${{B_s}}$ transmitted over V2V links is within the permissible limit $B_s^{\max }$, representing the upper bound of the semantic load capacity.
	
    In the proposed optimization problem, discrete variables such as binary channel allocation variables and continuous variables such as transmit power coexist, and nonlinearity is introduced by sigmoid functions. This makes it a mixed integer nonlinear programming (MINLP) problem. Furthermore, the dynamic network topology and time-varying channel conditions increase the complexity. These aspects make traditional optimization methods ineffective. To address the problem, we model it as a Markov decision process (MDP) suitable for DRL. DRL is well-suited for this context as it learns adaptive policies through interaction with the environment\cite{n4}. However, centralized DRL approaches suffer from high communication overhead and scalability issues. We therefore utilize multi-agent deep deterministic policy gradient (MADDPG), a decentralized learning framework that enables each vehicle (i.e., agent) to independently learn optimal strategies while considering the actions of others. This is essential for multi-agent systems like platooning. Additionally, MADDPG’s capacity to manage continuous action spaces aligns it well with the complex resource management tasks in V2X communication.
	
	\section{Proposed SAMRAMARL Algorithm Approach}
	To mitigate the computational complexity of the non-linearity and mixed-integer nature of the problem, we apply a series of simplifications. Specifically, the binary channel allocation variables ${{\beta _{n,k}}}$ are relaxed into continuous variables in the range $[0,1]$:
	\begin{equation}\label{P1h1} \beta _{n,k}^t \in [0,1], \quad \forall n \in \mathcal N, \forall k \in \mathcal K. \end{equation}
	This relaxation transforms the problem from a MINLP problem into a non-linear programming (NLP) problem, which is more amenable to continuous optimization techniques. After solving the relaxed problem, a thresholding method can be used to map the continuous values back to binary decisions, providing a practical solution to the original problem.
	Additionally, the objective function (\ref{P1a}), which is a weighted sum of the QoE term and the probability of timely delivery, is further simplified to facilitate computation. For SRS, we approximate it using a logistic function:
	\begin{equation}\label{P1h2} \frac{1}{{1 + \exp( - \alpha (\sum\limits_{t = 1}^T {\sum\limits_{m = 1}^{M_{n}} {{\beta _{n,k}}} } \phi _k[m,t] - \frac{{{B_s}}}{{\Delta T}}))}}, 
	\end{equation}
	where $\alpha$ is a scaling parameter that controls the sensitivity of the logistic function. This approximation allows the probability calculation to become continuous and differentiable, making it suitable for gradient-based optimization methods.
	With these simplifications in place, we turn to the multi-agent reinforcement learning approach to solve the problem efficiently.
	
	In this section, we will explore the multi-agent environment, detailing its states, actions, and rewards. We will then introduce the proposed SAMRAMARL algorithm and present its key formulations. Here, each PL interacts with the environment as an agent and makes decisions through the corresponding strategies. At time slot $t$, the agent obtains the current state $s_t$, the corresponding action $a_t$, and the corresponding reward $r_t$, and transitions to the next state $s_{t+1}$. This process can be formulated as ${e_t} = ({s_t},{a_t},{r_t},{s_{t + 1}})$.
	
	\subsection{Modeling of the System}
	Now, we first construct this DRL framework with states, actions and rewards and use SAMRAMARL algorithm to find the optimal policy. The relevant details are described below:
	
	\textbf{State space:} The state of the $n$th PL at time slot $t$ consists of the following components: the instantaneous channel information of the $n$th PL and the BS $h_{n, BS}^t[k]$, the instantaneous channel information of the $n$th PL and his followers $h_{n,m}^t[k]$, the previous interference from other platoons $I_{n,m}^t[k]$ and $I_{T,BS}^t[k]$, the residuals transmitted by the intra-platoon communication loads $B_{s,n}^t$, and thus, the state space can be expressed as:
	\begin{equation}
		s_n^t = \left[ h_{n, BS}^t[k], h_{n,m}^t[k], I_{n,m}^t[k], I_{T,BS}^t[k], B_{s,n}^t \right].
	\end{equation}
	
	\textbf{Action space:} As mentioned before, PL contains several actions: $\beta_{n,k}$ is used to select the sub-channel for communication, $\rho_{n,k}$ determines inter or intra communication, the power level of the communication $p_{T,n}[k]$ and $p_{I, n}[k]$, and the sentence lengths corresponding to V2I communication and V2V communication ($u_m^T$ and $u_m^I$). Note that when the PL selects inter communication, it has only one sentence length to select; when the PL selects intra communication, there will be $M$ followers of sentence lengths to select. Thus, the action space of agent $n$ at time slot $t$ is defined as:
	\begin{equation}
		a_n^t = \left[ \beta_{n,k}^t, \rho_{n,k}^t, p_{T,n}^t[k], p_{I, n}^t[k], u_m^{T,t}, u_m^{I,t} \right].
	\end{equation}
	
	\textbf{Reward function:} in our proposed scheme, there are two aspects of rewards for each agent, one is the global reward reflecting the cooperation among agents, and the other is the local reward that is to help each agent to explore the optimal action.
	In DRL, rewards can be set flexibly, and a good reward can improve the performance of the system. Here, our goal is to optimize the agent $n$'s QoE level, as well as the corresponding payload transmission probability. Thus the local and global rewards of the $n$th PL at time slot $t$ are:
	\begin{equation}\label{eq20}
		\begin{array}{l}
			{r_{n,l}^{t}} =  - {w_1}\frac{1}{{1 + \exp ( - \alpha (\sum\limits_{t = 1}^T {\sum\limits_{m = 1}^{M_{n}} {{\beta _{n,k}}} } \phi _k[m,t] - \frac{{{B_s}}}{{\Delta T}}))}}\\
			{\rm{        }} \qquad\qquad\qquad\qquad\qquad\qquad\quad+ {w_2}Qo{E^n},
		\end{array}
	\end{equation}
	and \begin{equation}\label{eq21}
		{r_{g}^{t}} = \frac{1}{N}\sum\limits_{n \in \mathcal N} {{r_{n,l}^{t}}}.
	\end{equation}
	
	\subsection{SAMRAMARL Algorithm}
	\begin{algorithm}
		\caption{SAMRAMARL Algorithm}\label{al2}
		Start environment simulator, generate vehicles\\
		Initialize global critic networks $Q_{{\psi _1}}^{{g_1}}$ and $Q_{{\psi _2}}^{{g_2}}$\\
		Initialize target global critic networks $Q_{{\psi _1'}}^{{g_1}}$ and $Q_{{\psi _2'}}^{{g_2}}$\\
		Initialize each agent's policy and critic networks\\
		\For{each episode}
		{
			Reset simulation paramaters\\
			\For{each timestep $t$}
			{
				
				\For{each agent $n$}
				{
					Observe $s_n^t$ and select action $a_n^t=\pi_\theta(s_n^t)$\\
				}
				${\bf{s}} = ({s_1},{s_2}, \cdots {s_P})$ and ${\bf{a}} = ({a_1},{a_2}, \cdots {a_P})$ \\
				Receive global reward $r_g^t$ and local reward $r_{n,l}^t$\\
				Store $({{\bf{s}}^t},{{\bf{s}}^t},{\bf{r}}_l^t,r_g^t,{{\bf{s}}^{t + 1}})$ in replay buffer $\mathcal{B}$\\
				
				\If{the replay buffer size is larger than ${{\cal B}_{th}}$}
				{
					Randomly sample mini-batch of ${{\cal B}_{th}}$ transitions tuples from $\mathcal{B}$\\
					Update global critics by minimizing the loss according to Eq.(\ref{eq18})\\
					Update global target networks parameters: ${\psi _j'} \leftarrow \tau {\psi _j} + (1 - \tau ){\psi _j'}, j=1,2$\\
					
					\If{episode mod $d$}
					{
						\For{each agent $k$}
						{
							Update local critics by minimizing the loss according to Eq.($\ref{eq20}$)\\
							Update local actors according to Eq.($\ref{eq17}$)\\
							Update local target networks parameters: ${\theta _n'} \leftarrow \tau {\theta _n} + (1 - \tau ){\theta _n'}$, ${\phi _n'} \leftarrow \tau {\phi _n} + (1 - \tau ){\phi _n'}$
						}
					}
				}
			}
		}	
	\end{algorithm}
	
	The SAMRAMARL algorithm we proposed is designed to optimize both global system objectives and individual agent behavior in a dynamic, semantic communication environment, particularly tailored for vehicular platooning scenarios. The algorithm employs a multi-agent actor-critic approach, where each agent is equipped with its own actor and local critic networks for decision-making, while a pair of global critics guides the overall system optimization \cite{RN8}.
	
	Specifically, we consider a vehicular environment with $N$ platoons and the policies for all agents are $\mathbf{\pi} = \{ {\pi _1},{\pi _2}, \cdots ,{\pi _N}\}$. The agent $n$'s strategy $\pi_n$, Q-functions 
	${\mathop{\rm Q}\nolimits} _{{\phi _n}}^k$ and twin global critic Q-functions 
	(${\mathop{\rm Q}\nolimits} _{\psi 1}^{{g_1}}$, ${\mathop{\rm Q}\nolimits} _{\psi 2}^{{g_2}}$) are parameterized by ${\theta _n}$, $\phi_n$, ${\psi _1}$ and $\psi_2$, respectively.
	The training process begins with an initial  the environment simulator, configured with $P$ vehicular platoons to reflect a realistic multi-agent communication scenario. Initially, twin global critic networks, $Q_{{\psi _1}}^{{g_1}}$ and $Q_{{\psi _2}}^{{g_2}}$ are initialized to evaluate joint state-action pairs across all agents. These global critics are responsible for assessing the performance of the entire system based on global rewards which is QOE.  
	Alongside, each agent $n$ initializes its policy network ${\pi _n}$	for action selection and a local critic network $Q_{{\phi _n}}^n$ to evaluate the quality of its actions based on local observations.
	
	At each time step, agent $n$ observes its local state $s_n^t$ and selects an action $a_n^t = {\pi _n}(s_n^t)$ . These actions of all agents form the action vector ${{\bf{a}}^t}$, while the states of all agents constitute the global state vector ${{\bf{s}}^t}$. The agents then receive a global reward $r_g^t$ that reflects the performance of the entire system, aiming to maximize metrics like overall communication efficiency and QoE. Additionally, each agent receives a local reward $r_{n,l}^t$ that is tailored to its specific semantic communication needs, allowing it to optimize its actions based on its individual context. All experiences $({{\bf{s}}^t},{{\bf{a}}^t},{\bf{r}}_l^t,r_g^t,{{\bf{s}}^{t + 1}})$ are stored in a shared replay buffer ${\cal B}$. This shared experience pool enables agents to learn not only from their own interactions but also from the collective experiences of all agents, facilitating more effective and stable learning through diverse sampling. When the buffer size reaches a predefined threshold ${{\cal B}_{th}}$, a mini-batch of transitions is randomly sampled from ${\cal B}$, facilitating the update of both global and local critics.
	The global critics $Q_{{\psi _1}}^{{g_1}}$ and $Q_{{\psi _2}}^{{g_2}}$ are updated by minimizing the loss function:
	\begin{equation}\label{eq18}
		L({\psi _j}) = {\rm{E}}\left[ {{{(Q_{{\psi _j}}^{{g_j}}({\bf{s}},{\bf{a}}) - {y_g})}^2}} \right],j = 1,2,
	\end{equation}
	where 
	\begin{equation}\label{eq19}
		{y_g} = {r_g} + \gamma {\mathop {\min }\limits_{j = 1,2}}{\left. {Q_{\psi _j'}^{{g_j}}({\bf{s',a'}})} \right|_{a_n' = \pi _n'(s_n')}}.
	\end{equation}
	In this context, $\gamma$ is the discount factor, ${\bf{\pi}}' = \{ \pi _1',\pi _2', \cdots ,\pi _N'\}$ is the target policy with parameter $\theta ' = \{ \theta _1',\theta _2', \cdots ,\theta _N'\}$. The use of twin critics mitigates the overestimation bias in value estimation, leading to more stable and accurate training.
	Similarly, the local critic $Q_{{\phi _n}}^n$ of agent $n$ is updated to 
	\begin{equation}\label{eq20}
		{L_n}({\phi _n}) = {\rm{E}}\left[ {{{(Q_{{\phi _n}}^k({s_n},{a_n}) - y_l^n)}^2}} \right],
	\end{equation}
	where\begin{equation}\label{eq1}
		y_l^n = r_l^n + \gamma Q_{{\phi _{n'}}}^k({s_{n'}},{\pi _{n'}}({s_{n'}})).
	\end{equation}
	
	The update of the agent’s policy ${\pi _n}$ combines the feedback from both global and local critics, enabling each agent to balance between system-wide and individual objectives.
	The policy gradient for agent $n$ is given by:
	\begin{equation}\label{eq17}
		\begin{array}{l}
			\nabla J({\theta _n}) = \left[ {{\nabla _{{\theta _n}}}{\pi _n}({a_n}|{s_n}){\nabla _{{a_n}}}Q_{{\psi _j}}^{{g_j}}({\bf{s}},{\bf{a}})} \right] + \\
			{\rm{                }}\qquad\qquad\left[ {{\nabla _{{\theta _n}}}{\pi _n}({a_n}|{s_n}){\nabla _{{a_n}}}Q_{{\phi _n}}^n({s_n},{a_n})} \right],
		\end{array}
	\end{equation}
	where ${\bf{s}} = ({s_1},{s_2}, \cdots {s_P})$ and 
	${\bf{a}} = ({a_1},{a_2}, \cdots {a_P})$ are the total state and action vectors.
	The first term in this gradient involves the global critic, guiding the agent to actions that improve the overall system performance. The second term involves the local critic, which adjusts the agent's actions to better align with its local observations and specific communication needs.
	
	To ensure that the learned policies and value estimates remain stable during training, target networks are used for both the global critics and the local actor-critic pairs. These target networks are updated through a soft update mechanism:
	\begin{equation}\label{eq29}
		{\theta _{n'}} \leftarrow \tau {\theta _n} + (1 - \tau ){\theta _{n'}}
	\end{equation}
	\begin{equation}\label{eq30}
		{\phi _{n'}} \leftarrow \tau {\phi _n} + (1 - \tau ){\phi _{n'}}
	\end{equation}
	where $ \tau  \in (0,1)$ is a smoothing factor that controls the rate at which the target networks follow the main networks. This approach reduces the variance of target values during updates, further stabilizing the learning process.
	
	The detailed SAMRAMARL algorithm is described in the Algorithm \ref{al2}.	
	Since the algorithm is implemented as the actor-critic framework, denote the computational complexities for calculating gradients and updating parameters in the actor and critic networks being denoted as $G_A$, $G_C$, $U_A$, and $U_C$ respectively. Since the architecture of the target actor network and the target critic network is the same as the network structure of the two mentioned above, their complexities remain the same. Thus, the computational complexity of the SAMRAMARL algorithm is calculated as
	$\mathcal{O}\left( {{G_A} + {G_C} + 2{U_A} + 2{U_C}} \right) + {\text{ }}\mathcal{O}\left( {2{G_C} + 4{U_C}} \right)$.
	
	\section{Simulation Results}
	In this section, We have evaluated the performance of the proposed SAMRAMARL algorithm in terms of the QoE and SRS.
	\subsection{Simulation Settings and Dataset}
	We simulate a cellular-based vehicular communication network situated in an urban environment in Fig. \ref{fig1}. The simulation is conducted using Python 3.9, with the communication settings adhering to both V2V and V2I communication models, based on the architecture specified by 3GPP TR 36.885 \cite{RN90}. The simulation explores variations in intra-platoon spacing, semantic demand sizes, and the average semantic symbol length ${u_m}$, which significantly impact the overall QoE and SRS for V2V transmissions.
	Each road consists of four lanes in two directions, with a lane width of 3.5 meters. The distance between adjacent intersections ${d_{{\rm{adj}}}}$ is 433 meters. 
	Throughout the simulation, all lanes maintain uniform vehicle density and speed, traveling at a constant speed of 36 kilometers per hour. 
	The minimum dimensions of the simulated area are 1299 meters by 750 meters.
	We assume that large-scale fading remains constant throughout each episode, while small-scale fading evolves dynamically. Key vehicular environment and neural network parameters are outlined below. Detailed parameter settings are listed in Table~\ref{tab:vehicular_parameters}.
	\begin{table}[t]
		\setlength{\tabcolsep}{-0.2pt}
		\begin{center}
			\caption{Vehicular Environment Parameters}
			\begin{tabular}{|c|c|}
				\hline
				\textbf{Vehicular Environment Parameters} & \textbf{Value} \\ \hline
				Carrier frequency & 2 GHz \\ \hline
				Number of resource blocks (RBs) & 4 \\ \hline
				Bandwidth per RB & 180 kHz \\ \hline
				Number of vehicles & 20 \\ \hline
				Platoon size & 5 vehicles \\ \hline
				Platoon speed & 36 km/h \\ \hline
				Intra-platoon gap & 5-35 m \\ \hline
				Antenna height (RSU/Vehicle) & 25 m / 1.5 m \\ \hline
				Antenna gain (RSU/Vehicle) & 8 dBi / 3 dBi \\ \hline
				Receiver noise figure (RSU/Vehicle) & 5 dB / 9 dB \\ \hline
				Maximum vehicle transmit power & 30 dBm \\ \hline
				Noise power (\( \sigma^2 \)) & -114 dBm \\ \hline
				Semantic demand size & 1000-6000 / \( u_q \) symbols \\ \hline
				Average semantic symbol length \( u_q \) & 5-30 symbols \\ \hline
				Path loss model (V2I/V2V links) & 128.1 + 37.6 log10(d) \\ \hline
				Shadowing distribution & Log-normal \\ \hline
				\makecell{Shadowing standard deviation \\(V2I/V2V links)}  & 8 dB / 3 dB \\ \hline
				Decorrelation distance (V2I/V2V links) & 50 m / 10 m \\ \hline
				Pathloss/shadowing update (V2I/V2V) & Every 100 ms \\ \hline
				Fast fading update (V2I/V2V) & Every 1 ms \\ \hline
				Fast fading model & Rayleigh fading \\ \hline
				QoE weight \( \omega_q \) & \( \sim U(0, 1) \) \\ \hline
				Required semantic accuracy \( {\xi _{target}} \) & \( \sim U(0.8, 0.9) \) \\ \hline
				Semantic rate requirement for text \( {\phi _{target}} \) & \( \sim U(50, 70) \) ksuts/s \\ \hline
				Semantic rate requirement for images \( {\phi _{target}} \) & \( \sim U(80, 100) \) ksuts/s \\ \hline
				QoE threshold \( G_{\text{th}} \) & 0.5 \\ \hline
				Growth rate of $Scor{e_{\rm{R}}}$ \( {\gamma _q} \) & \( \sim N(0.1, 0.02^2) \) \\ \hline
				Growth rate of $Scor{e_A}$ \( {\delta _\mu } \) & \( \sim N(55, 2.5^2) \) \\ \hline
			\end{tabular}
			\label{tab:vehicular_parameters}
		\end{center}
	\end{table}
	
	The single-modal text dataset selected for extracting semantic information from text is the European Parliament dataset \cite{RN91}, consisting of approximately 2.0 million sentences and 53 million words.
	The dataset is pre-processed, and sentence lengths are adjusted to range from 1 to 20 words. The first 90\% of the dataset is used for training, and the remaining 10\% is for testing.
	The multi-modal text and image dataset is CLEVR \cite{CLEVR}, which consists of a
	training set of 70,000 images and 699,989 questions and a test
	set of 15,000 images and 149,991 questions.
	The settings for SAC parameters are shown in Table~\ref{tab:neural_network_parameters}.
	\begin{table}[htbp]
		\centering
		\caption{Neural Network Parameters}
		\begin{tabular}{|l|l|}
			\hline
			\textbf{Neural Network Parameters} & \textbf{Value} \\ \hline
			Replay buffer size & 1000,000 \\ \hline
			Mini-batch size & 64 \\ \hline
			Local actor network layers (number/size) & 2 / (1024, 512) \\ \hline
			Local critic network layers (number/size) & 3 / (1024, 512, 256) \\ \hline
			Global critic network layers (number/size) & 3 / (1024, 512, 256) \\ \hline
			Critic/Actor learning rate & 0.001 / 0.0001 \\ \hline
			Discount factor & 0.99 \\ \hline
			Soft update parameter (\( \tau \)) & 0.005 \\ \hline
			Number of episodes & 500 \\ \hline
			Iterations per episode & 100 \\ \hline
			Policy update delay factor & 2 \\ \hline
			Gaussian noise distribution & \( \mathcal{N}(0, 0.2) \) \\ \hline
		\end{tabular}
		\label{tab:neural_network_parameters}
	\end{table}
	\vspace{-0.8cm}
	\subsection{Performance Evaluation}
	
	To assess the performance of the proposed distributed semantic-aware multi-modal resource allocation framework (SAMRAMARL), we utilize QoE and SRS in V2X links as performance metrics. Furthermore, communication delay is also considered, since it can directly reflect the real-time performance of semantic information transmission. We compare SAMRAMARL against the following baseline methods:
	\begin{itemize} 
		\item \textbf{DDPG\_NO\_SC:} A DDPG-based algorithm that does not consider semantic information for resource allocation. This method is referred to as $DDPG\_NO\_SC$ in our comparisons.
		
		\item \textbf{DDPG:} A DDPG-based algorithm that incorporates semantic information into resource allocation, denoted as $DDPG$.
		
		\item \textbf{TD3:} A twin delayed deep deterministic policy gradient (TD3)-based algorithm that also utilizes semantic information. This method is referred to as $TD3$.
	\end{itemize}
	
	In traditional communication systems, semantic information is transmitted using bits rather than symbols explicitly designed for semantic representation. Although each bit can be loosely considered a semantic symbol, it generally conveys less semantic information compared to symbols employed in semantic transmission methods like DeepSC. Therefore, in order to compare with semantic communication, $QOE'$ is used in traditional communication, defined similar to equation (\ref{eq66}):
	\begin{equation}\label{eq31}
		QOE' = \frac{{W{{\tilde H}_{SM/MM}}}}{{u_m^{T/I}}},
	\end{equation}
	where ${u_m^{T/I}}$ is defined as the transform factor, with unit of bits/word. For example, if a word consists of 5 letters and ASCII code is used for encoding, then ${u_m^{T/I}}$ is 40 bits/word.
	We assume that ${{\tilde H}_{SM/MM}}$ is fixed, indicating no bit errors in conventional communications.
	
	\begin{figure}[htbp]
		\centering
		\includegraphics[width=0.47\textwidth]{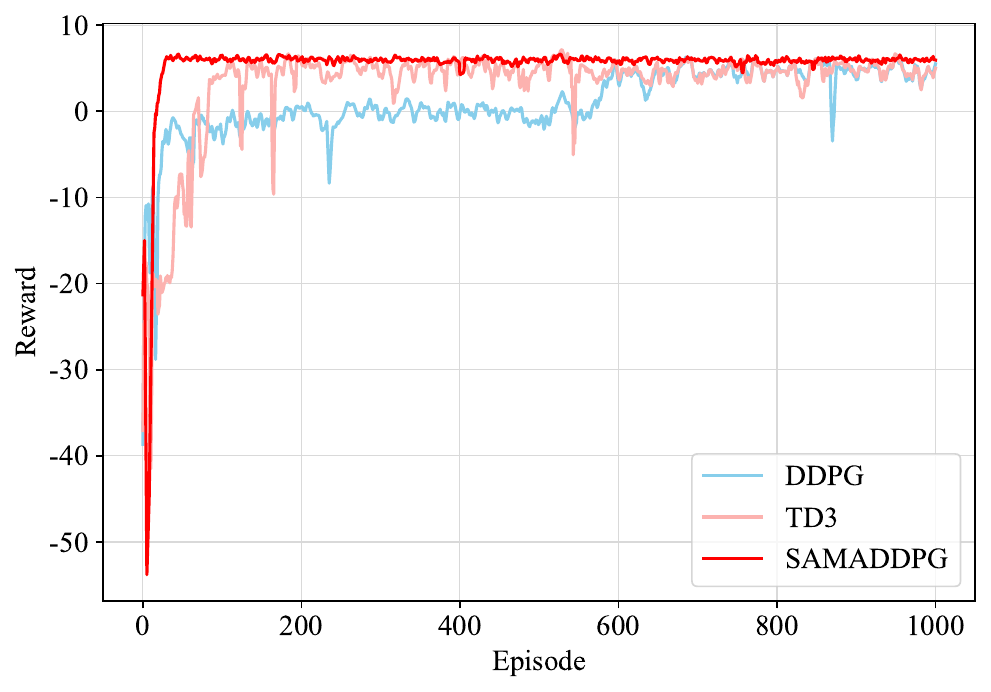}
		\caption{Training rewards across different resource allocation algorithms}
		\label{fig_rewards}
		\vspace{-0.2cm}
	\end{figure}
	
	In Fig. \ref{fig_rewards}, we analyze the reward evolution of various algorithms, including our SAMRAMARL framework. SAMRAMARL exhibits faster convergence and greater stability than $DDPG$ and $TD3$, attributable to its multi-agent structure. This decentralized decision-making allows each vehicle to respond independently to local conditions, optimizing resource allocation in dynamic environments. While $DDPG$ and $TD3$ also consider semantic awareness, their single-agent models struggle to adapt quickly to environmental changes. Specifically, $DDPG$ faces instability from a poor exploration-exploitation balance, resulting in fluctuating rewards, while $TD3$ converges more slowly than SAMRAMARL, highlighting the benefits of multi-agent systems.
	The distributed nature of SAMRAMARL fosters better collaboration, leading to quicker convergence and stability, particularly under rapidly changing semantic information. This approach effectively meets the varying demands of vehicles within a platoon, ensuring efficient spectrum sharing and resource allocation.
	
	\begin{figure}[htbp]
		\centering
		\includegraphics[width=0.47\textwidth]{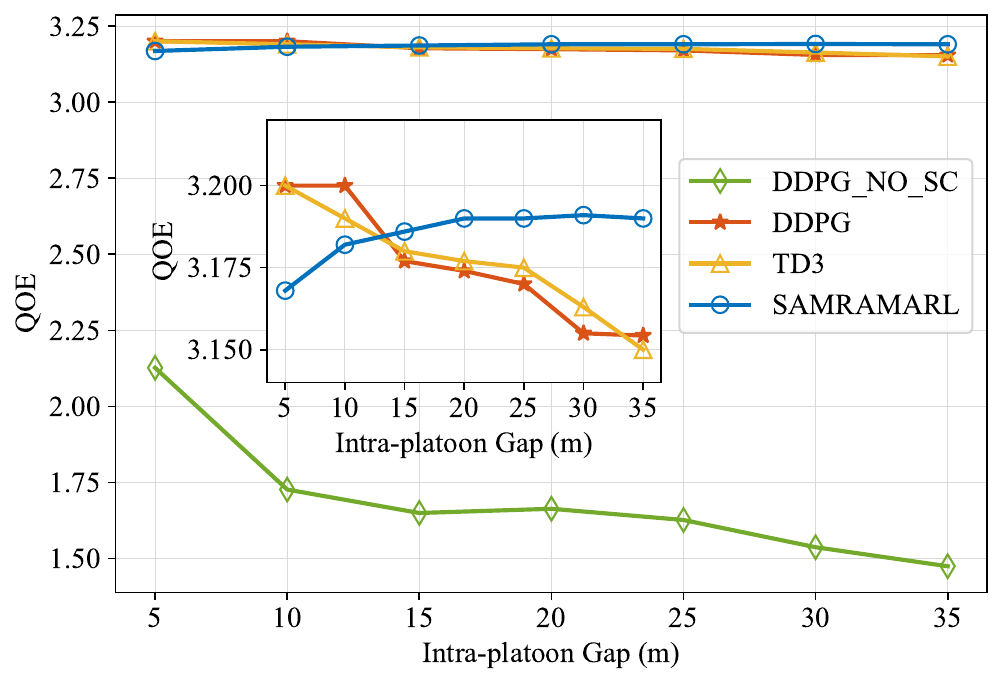}
		\caption{Intra platoon Gap v.s. QOE with semantic demand size $= 4$ ksut.}
		\label{fig3}
		\vspace{-0.2cm}
	\end{figure}
	
	Fig. \ref{fig3} illustrates how the intra-platoon gap, or the physical distance between vehicles, impacts QoE. As this gap increases, communication quality typically deteriorates, leading to a drop in QoE for traditional algorithms like $DDPG$ and $TD3$. However, SAMRAMARL exhibits a unique pattern: its QoE initially improves with increasing gaps due to its ability to dynamically adjust semantic symbol lengths and power allocation, reducing interference. This capability enables SAMRAMARL to maintain consistent QoE even as the gap widens, in contrast to the continuous decline seen in $DDPG$ and $TD3$.
	\begin{figure}[htbp]
		\centering
		\includegraphics[width=0.47\textwidth]{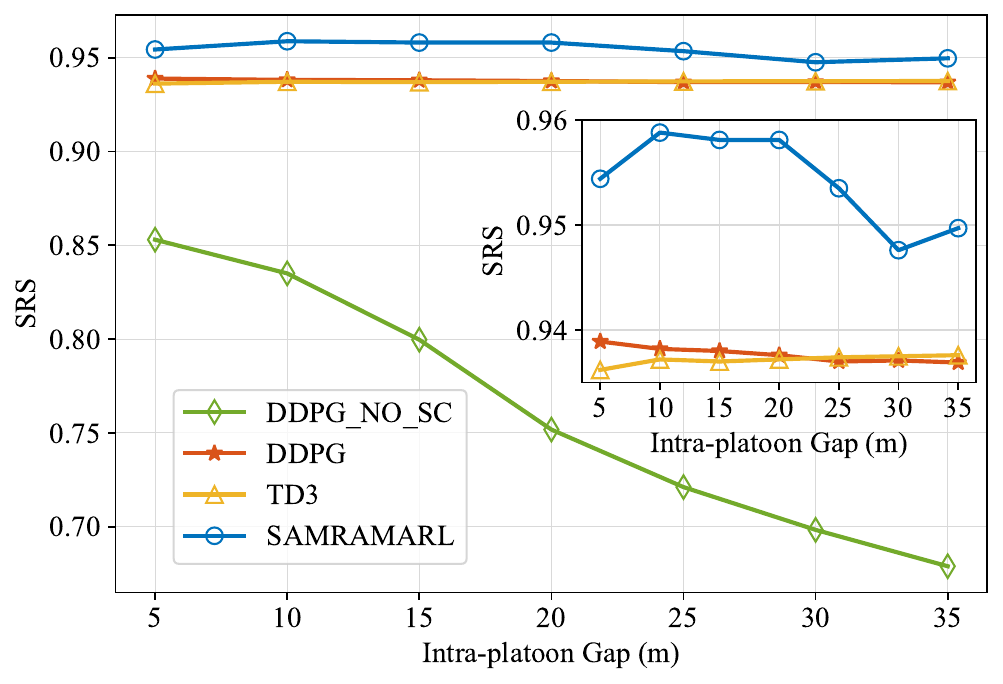}
		\caption{Intra platoon Gap v.s. SRS with semantic demand size $= 4$ ksuts.}
		\label{fig4}
		\vspace{-0.2cm}
	\end{figure}
	
	Fig. \ref{fig4} presents the relationship between intra-platoon gap and SRS. The $DDPG\_NO\_SC$ curve shows a clear decline with increasing gaps, reflecting increased signal attenuation. In contrast, $DDPG$ and $TD3$ maintain relatively stable SRS values, indicating their effectiveness in leveraging semantic information. Notably, SAMRAMARL outperforms both traditional algorithms, showcasing its advanced resource allocation strategies.
	
	\begin{figure}[htbp]
		\centering
		\includegraphics[width=0.47\textwidth]{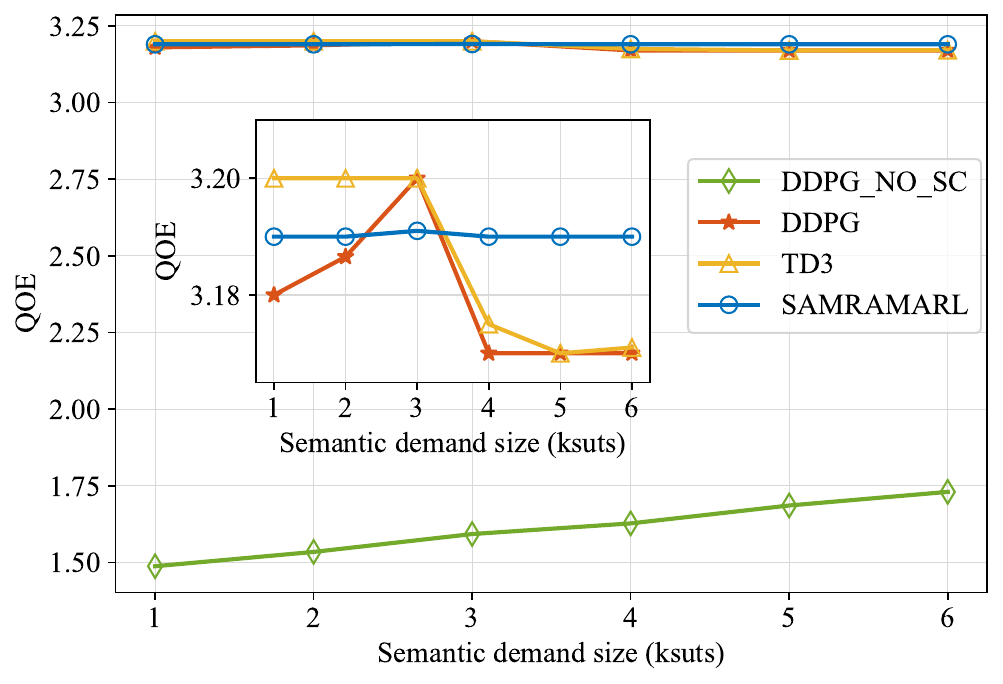}
		\caption{semantic demand size v.s. QOE with intra platoon gap $= 20$ m.}
		\label{fig5}
		\vspace{-0.2cm}
	\end{figure}
	
	Fig. \ref{fig5} illustrates the impact of semantic demand size on the QoE.
	As the size of the semantic data packets increases, the QoE for all algorithms improves, reflecting the direct relationship between semantic demand size and QoE. Larger semantic data packets allow more comprehensive information transmission, enhancing overall communication effectiveness.
	However, at a semantic demand size of 2000 suts, SAMRAMARL achieves a QoE level that is relatively mid-range among the three methods. 
	At a demand size of 3000 suts, we observe a decline in QoE for $DDPG$ and $TD3$, indicating their struggle to adapt to the increased data load. In contrast, SAMRAMARL retains stable performance, which can be attributed to its robust reinforcement learning framework. 
	
	\begin{figure}[htbp]
		\centering
		\includegraphics[width=0.47\textwidth]{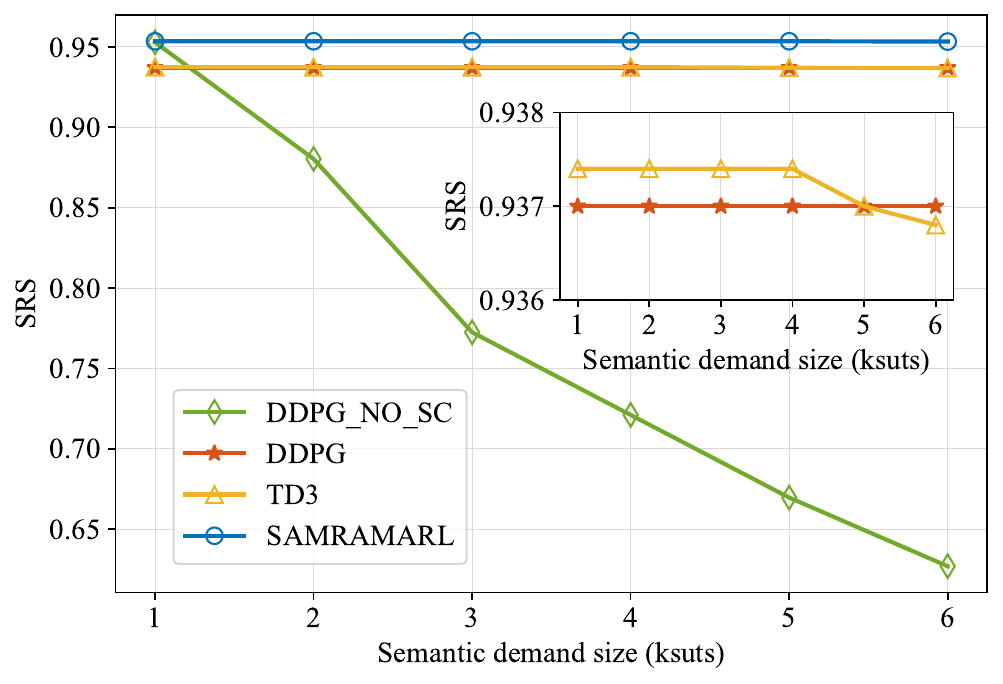}
		\caption{Semantic demand size v.s. SRS with intra platoon gap $= 20$ m.}
		\label{fig6}
		\vspace{-0.2cm}
	\end{figure}
	
	Fig. \ref{fig6} examines the relationship between semantic demand size and SRS. The curve for $DDPG\_NO\_SC$ declines linearly as demand size increases, highlighting its inability to manage the growing volume of semantic data. In contrast, $DDPG$ and $TD3$ show stability, effectively utilizing semantic context to maintain performance. SAMRAMARL consistently achieves higher SRS, attributed to its multi-agent reinforcement learning framework, which facilitates resource distribution and adapts to vehicular environment nuances.
	
	\begin{figure}[htbp]
		\centering
		\includegraphics[width=0.47\textwidth]{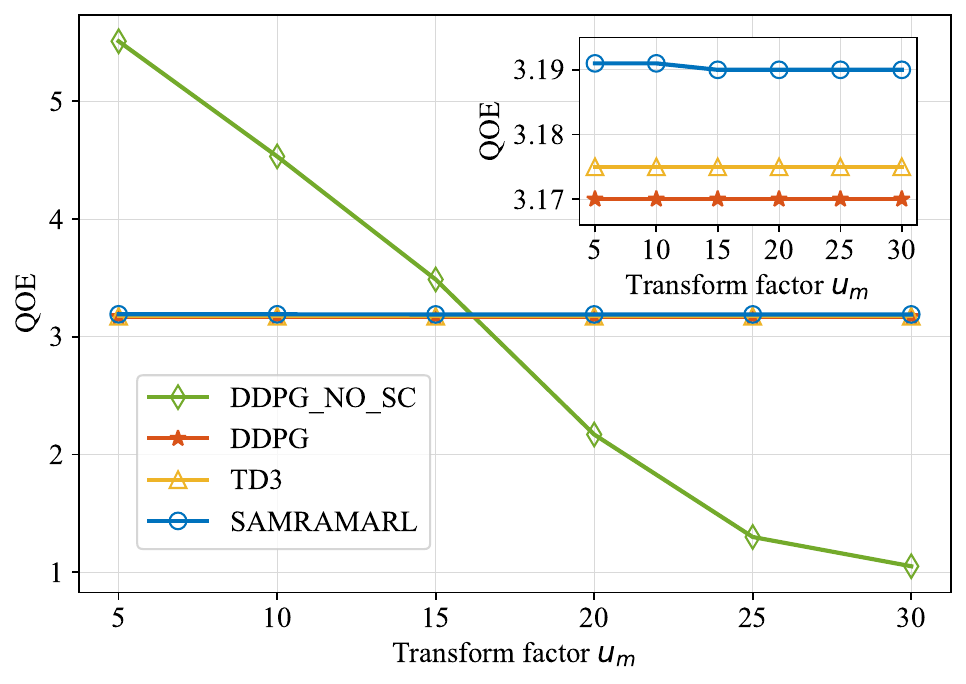}
		\caption{Transform factor ${u_m}$ v.s. QOE with semantic demand size $= 4$ ksut and intra platoon gap $= 20$ m.}
		\label{fig7}
		\vspace{-0.2cm}
	\end{figure}
	
	Fig. \ref{fig7} shows the relationship between the transform factor ${u_m}$ and QoE. The $DDPG\_NO\_SC$ curve declines with increasing ${u_m}$, reflecting the traditional QoE calculation's sensitivity to complexity. However, semantic-aware algorithms maintain stable QoE levels, as they focus on the relevance of transmitted information rather than just symbol length. Among these, SAMRAMARL achieves the highest QoE due to its dynamic resource allocation capabilities.
	
	\begin{figure}[htbp]
		\centering
		\includegraphics[width=0.47\textwidth]{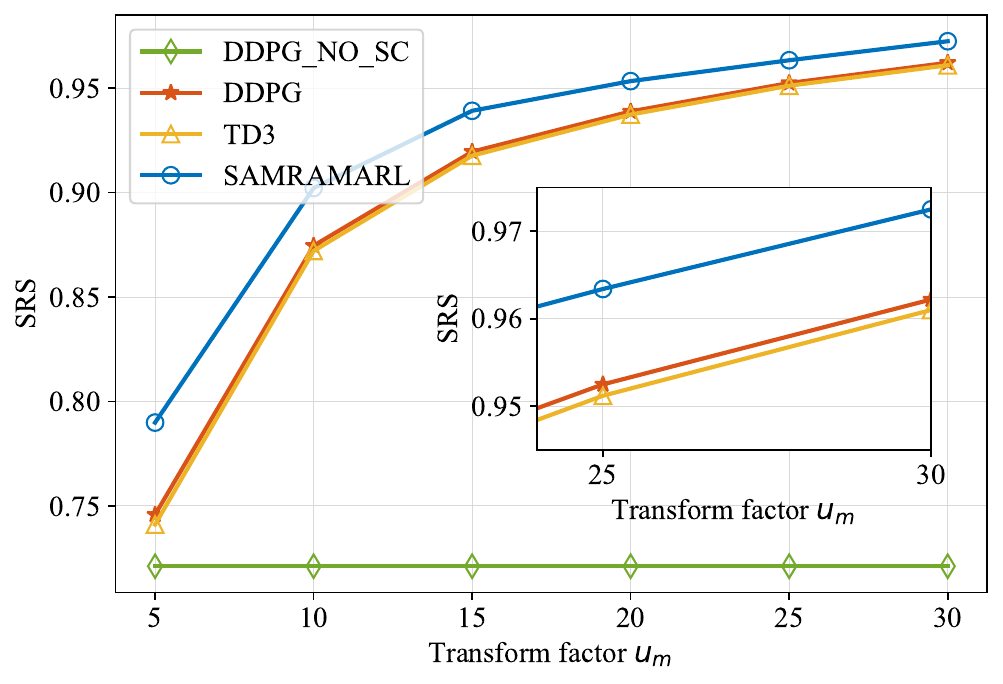}
		\caption{Transform factor ${u_m}$ v.s. SRS with semantic demand size $= 4$ ksut and intra platoon gap $= 20$ m.}
		\label{fig8}
		\vspace{-0.2cm}
	\end{figure}
	
	Fig. \ref{fig8} presents the relationship between the transform factor 
	${u_m}$ and SRS. The curve for the algorithm without semantic awareness remains constant as ${u_m}$ increases. Conversely, the curves for the semantic-aware algorithms demonstrate an upward trend with increasing ${u_m}$. This increase can be attributed to the fact that the semantic-aware algorithms optimize the communication process by prioritizing the transmission of relevant semantic content, thus enhancing the effectiveness of data transfer as ${u_m}$ grows. Our proposed SAMRAMARL algorithm achieves the highest SRS among the three, reflecting its ability to effectively allocate resources and adapt to dynamic network conditions.
	
	\begin{figure}[htbp]
		\centering
		\includegraphics[width=0.47\textwidth]{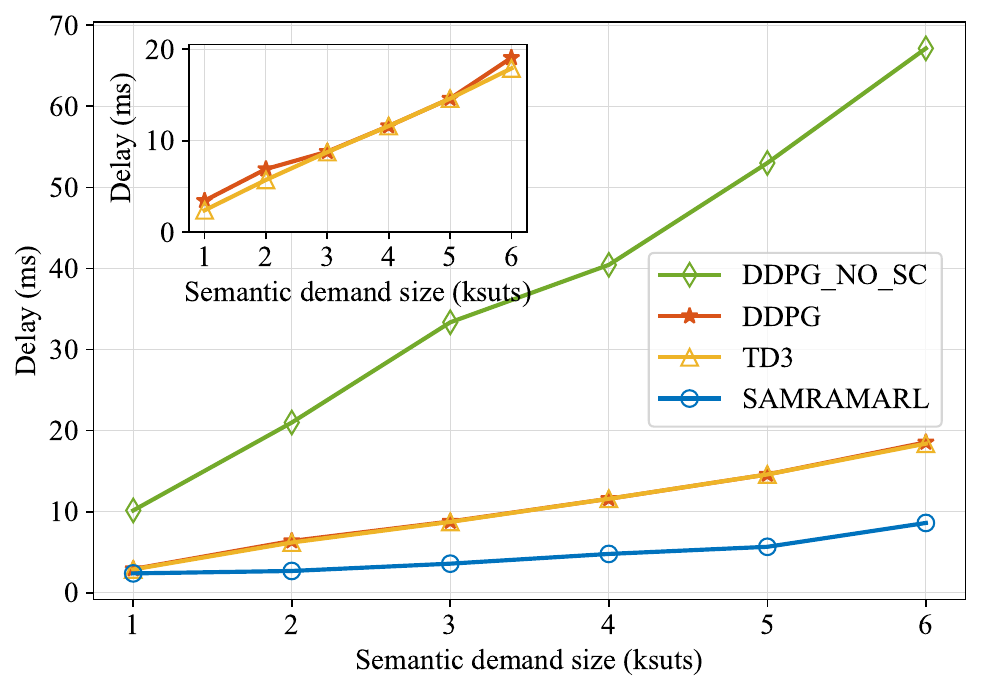}
		\caption{Semantic demand size ${u_m}$ v.s. Delay with intra platoon gap $= 20$ m.}
		\label{fig9}
		\vspace{-0.2cm}
	\end{figure}
	
	In Fig. \ref{fig9}, the communication delay increases for all methods as the semantic demand size increases, but communication delay of SAMRAMARL is always the lowest and has the smallest growth rate. This is attributed to the multi-agent reinforcement learning framework, which can efficiently manage the transmission of semantic information and significantly enhance the robustness of the system. $DDPG\_NO\_SC$ has a relatively large delay because it cannot perceive semantics. Moreover, with the increase of semantic demand size, the delay of $DDPG\_NO\_SC$ shows an obvious growth trend, indicating that it cannot effectively cope with the load pressure brought by the increase of semantic demand size. $DDPG$ and $TD3$ are superior to $DDPG\_NO\_SC$, because they can sense semantics and significantly reduce transmission redundancy, thereby reducing delay. The performance of $TD3$ in terms of delay is similar to that of $DDPG$, but the overall delay is slightly lower due to the double-delay mechanism of $TD3$.
	
	\section{Conclusion}
	In this paper, we introduced a SAMRAMARL framework for platooning systems, leveraging MARL to enhance QoE and SRS in C-V2X environments. Simulation results have demonstrated the superior performance of SAMRAMARL compared to traditional resource allocation methods.
	
	The main results can be summarized as follows:
	\begin{itemize} 
		\item The incorporation of semantic communication improves the efficiency of resource management, leading to enhanced QoE and SRS in dynamic platooning scenarios. 
		\item Our proposed SAMRAMARL algorithm balances resource allocation among vehicles, enabling decentralized decision-making and adaptability to changing network conditions. 
		\item While QoE initially increases with semantic demand sizes, a threshold is identified beyond which QoE stabilizes, emphasizing the need for effective semantic feature information handling to avoid degradation in performance. 
	\end{itemize}
	The algorithm's ability to optimize channel assignment, power allocation, and the length of transmitted semantic symbols has demonstrated its robustness and potential for real-time applications in intelligent transportation systems.

	\nocite{*}
	\bibliographystyle{IEEEtran}
	\bibliography{ref}		
	
\end{document}